\documentclass[sigconf,screen,authorversion,nonacm,review=false,timestamp=false]{acmart}

\usepackage{hyperref}
\usepackage{amsthm}
\usepackage{subcaption}
\usepackage{cleveref}
\usepackage{multirow}

\usepackage{flushend}
\usepackage{balance}
\usepackage{xspace}
\usepackage{xcolor}

\usepackage{stfloats}
\usepackage{tikz}

\usepackage{amsmath, amsfonts}

\usetikzlibrary{shapes,fit} 
\usetikzlibrary{calc,trees}
\usetikzlibrary{arrows.meta}

\newcommand{\algoName}{\textsc{TAMLEC}\xspace}
\newcommand{\algoAblation}{$\sqrt{\textsc{TAMLEC}}$\xspace}

\newcommand\low{\text{low$_T$}}

\AtBeginDocument{%
 }

\makeatother

\begin{document}

\title{Extreme Multi-label Completion for Semantic Document Labelling with Taxonomy-Aware Parallel Learning}

 \author{Julien Audiffren}
 \affiliation{%
   \institution{University of Fribourg}
   \city{Fribourg}
   \country{Switzerland}}
 \email{julien.audiffren@unifr.ch}

  \author{Christophe Broillet}
 \affiliation{%
 	\institution{University of Fribourg}
 	\city{Fribourg}
 	\country{Switzerland}}

 \author{Ljiljana Dolamic}
 \affiliation{%
   \institution{armasuisse S+T}
   \city{Thun}
   \country{Switzerland}}
 \email{ljiljana.dolamic@ar.admin.ch}

 \author{Philippe Cudr{\'e}-Mauroux}
 \affiliation{%
  \institution{University of Fribourg}
   \city{Fribourg}
   \country{Switzerland}}
 \email{pcm@unifr.ch}

\begin{abstract}
In Extreme Multi Label Completion (XMLCo), the objective is to predict the missing labels of a collection of documents.
Together with XML Classification, XMLCo is arguably one of the most challenging document classification tasks, as the very high number of labels (at least ten of thousands) is generally very large compared to the number of available labelled documents in the training dataset.
Such a task is often accompanied by a taxonomy that encodes the labels organic relationships, and many methods have been proposed to leverage this hierarchy to improve the results of XMLCo algorithms.   
In this paper, we propose a new approach to this problem, \algoName (Taxonomy-Aware Multi-task Learning for Extreme multi-label Completion).
\algoName divides the problem into several Taxonomy-Aware Tasks, i.e. subsets of labels  adapted  to the hierarchical paths of the taxonomy, and trains on these tasks using a dynamic Parallel Feature sharing approach, where some parts of the model are shared between tasks while others are task-specific.
Then, at inference time, \algoName uses the labels available in a document to infer the appropriate tasks and to predict missing labels.
To achieve this result, \algoName uses a modified transformer architecture that predicts ordered sequences of labels on a Weak-Semilattice structure that is naturally induced by the tasks.
This approach yields multiple advantages.
First, our experiments on real-world datasets show that \algoName outperforms  state-of-the-art methods for various XMLCo problems.
Second, \algoName is by construction particularly suited for few-shots XML tasks, where new tasks or labels are introduced with only few examples, and extensive evaluations highlight its strong performance compared to existing methods.
\end{abstract}


\maketitle
\footnotetext[1]{Preprint version}

\newpage
\section{Introduction} \label{sec:intro}

In the recent decades, the number of textual documents (documents for short hereafter) available online has increased dramatically \cite{dong2017century}.  
This rise of large document collections has been further amplified in the last few years by Large Language Models and their ever-increasing need for new data corpora \cite{xue_repeat_2023}.
As a result, the automated labelling of documents has become a crucial issue \cite{DBLP:conf/nips/YouZWDMZ19}, as these labels permit documents categorization, help with the search of information and eases user navigation among vast collection of documents \cite{DBLP:journals/qss/WangSHWDK20}.
In particular, the semantic annotation of textual documents on the Web is an important application, as these labels are key to improve the search and discovery of relevant documents online, both for search engines and for users. 

%
%

\indent The problem of assigning to each document a subset of labels is referred to as Extreme Multi-Label Classification (XMLC)~\cite{DBLP:conf/nips/YouZWDMZ19}.
XMLC is arguably one of the most challenging document classification tasks \cite{liu_deep_2017}. 
First, the set of labels to choose from is very large, often numbering in the tens or hundred of thousands, and each document may possess an arbitrary number of labels (the \textit{labels scale} problem).
Second, the data is often considered sparse, in the sense that many labels only have few training instances, and the problem is further compounded when considering collections of labels (the \textit{scarcity} problem).
Put together, these issues make the use of traditional classification algorithms difficult and has given rise to new approaches to solve this problem.
A sub-problem of XMLC that is of particular interest to this work is  Extreme Multi-Label Completion (XMLCo) \cite{ostapuk2024follow}.
In this setting, each document is equipped with an incomplete set of labels, and the task is to predict the missing labels.
Compared to XMLC, XMLCo methods are able to leverage the known labels to deduce additional information about the document and improve their predictions.
The problem of incomplete labels is common in many applications \cite{DBLP:journals/cbm/RomeroNFRV23}, and can be due to  time constraints, subjectivity of the annotators, or the introduction of new labels in a dataset over time. As such, label completion is key to improving the quality of existing datasets \cite{DBLP:conf/kdd/PrabhuV14}.

\indent Some of the most successful methods proposed in the past few years have stemmed from the use of the label taxonomy \cite{DBLP:conf/www/ZhangSDW021,ostapuk2024follow,DBLP:journals/jodl/CaledSMW22}.  
Indeed, many real-world XMLC problems come equipped with a hierarchical label taxonomy, which is developed  to facilitate the management of large collections of labels.
These hierarchies encode the relation between labels that arise from their real-world usage, for instance through subsumption hierarchies specifying that a label (e.g., Philosophy) defines a more general class than a second label (e.g. Epistemology).
As a result, these structures yield valuable information for XMLC tasks, and in particular XMLCo, as in this case it has been observed that general, high labels are present while more specific labels are often missing: data instances are typically equipped with general, high level
labels, while more specific labels are more often missing \cite{DBLP:journals/qss/WangSHWDK20}.

For instance, scientific documents generally contain keywords that are parts of well-organized hierarchical ontologies, which naturally  yield label hierarchies \cite{DBLP:conf/www/ZhangSDW021}.
Notable XMLC related taxonomies include the MeSH thesaurus\footnote{https://www.nlm.nih.gov/mesh/meshhome.html} (for biomedical and health-related applications) and the Microsoft Academic Graph (MAG) \cite{DBLP:journals/qss/WangSHWDK20} (for scientific papers).
These taxonomies can be used to improve XMLC methods in multiple ways, for instance by acting as external regularization for label embedding \cite{DBLP:conf/www/ZhangSDW021} or by providing a natural clustering structure for the labels \cite{DBLP:journals/asc/GargiuloSCP19}. 
More recently, \cite{ostapuk2024follow} proposed a new XMLCo approach that uses a modified transformer architecture to directly perform prediction on the taxonomy, by generating paths of labels (from the more general to the more specific) on the tree that encodes the label hierarchy.
While these works have shown promising result and have highlighted the importance of using the taxonomy to solve the  label scale  problem, we argue that further performance can be gained by also using taxonomies to alleviate the scarcity problem.

\indent Multi-Task Learning (MTL) is a Machine Learning paradigm that involves training a model on multiple tasks simultaneously, with the goal of improving the performance of each task \cite{caruana1997multitask}. 
At the heart of the MTL challenge is also data scarcity, i.e. each task contains limited data, and training independent models on each task may result in poor performance and overfitting.
Thus algorithms that achieve good results in MTL generally do so by leveraging shared information and representations between tasks \cite{zhang2018overview}. 
One of the most popular approaches to achieve this is Parallel Feature Sharing, where multiple tasks are trained simultaneously by sharing a common feature extractor while maintaining task-specific components \cite{zhang2021survey}.
MTL and Parallel Feature Sharing have seen many applications in Natural Language Processing (NLP) and document analysis, and have received increased interest since the popularization of deep learning methods \cite{chen2024multi}.
Example of successful applications include sentiment analysis \cite{tan2023sentiment}, entity recognition \cite{aguilar2017multi} and question answering \cite{deng2019multi}. 

\indent The idea behind Parallel Feature Sharing can be found in many deep-learning XML algorithms such as AttentionXML \cite{DBLP:conf/nips/YouZWDMZ19}. 
Indeed, in most architectures the documents are processed through a shared neural network architecture, and only differentiated in the final result, which is generally a vector containing the predicted relevance of each label for the input document.
%
%
In that regard, only the weights of the last layers may be considered partially label-specific. 
However, and to the best of our knowledge, the use of more advanced MTL methods and in particular more intertwined shared/task-specific architectures and training methods have not received the same level of attention. 
Specifically, we argue that adapting the Parallel Feature Sharing paradigm to the taxonomy structure of the labels may lead to further improvements.

\indent  Building on this intuition, we introduce in this paper \algoName (Taxonomy-Aware Multitask Learning for Extreme multi-label Completion), which uses ideas from XMLCo and  MLT to better combine information sharing and the label taxonomy.
To achieve this, \algoName first creates Taxonomy-Aware Tasks (TATs), i.e. subsets of labels that are  adapted to the taxonomy structure and to semantic constraints (see Section \ref{subsec:semilattice}). 
\algoName uses a modified transformer architecture inspired by \cite{ostapuk2024follow} that is adapted to the characteristics of the TATs by balancing shared neurons and task-specific neurons to improve prediction performance. 
To predict missing labels, \algoName uses the known labels of each document to choose their relevant TATs,  and predict paths of labels on each of the selected task, which are then combined to perform the final prediction.
Our experiments on real-world datasets show that \algoName outperforms  state-of-the-art methods on various XMLCo problems.
Furthermore, we show that \algoName is by construction particularly suited for handling new tasks (few-shots XML tasks), where the new labels that form a TATs  are introduced after training with only few examples, and extensive evaluations highlight its strong performance compared to  existing methods.

The resulting neural network is trained using a new loss that embeds the different complexities and constraints of the different TATs.
 Our contributions can be summarized as follows: 
\begin{itemize}
	\item We extend the previous works on tree-based taxonomy  such as \cite{ostapuk2024follow} to accommodate a more general structure, Weak Semilattices, that naturally arises in many XML problems.
	\item We define the notion of Taxonomy-Aware Tasks (TATs), where the tasks are defined as sub-Weak-Semilattices of the taxonomy, that satisfy  coherence and separability criteria. These properties naturally result from the intuitive definition of tasks in XML, such as all the labels related to a broad category in the taxonomy (e.g. NLP-related labels as a subset of all Computer Science labels).
	\item We introduce a new algorithm, \algoName, that leverages an adaptive transformer architecture with a new TATs-aware loss to  balance information sharing among tasks for XMLCo problems.
	\item We perform an extensive empirical evaluation of our method, and show that it outperforms the state-of-the-art on XMLCo problems and performs competitively on XML Few-Shot tasks.
\end{itemize}
The rest of the paper is organized as follows. Section~\ref{sec:related} summarizes the related work in XML and MTL. We present our different contributions (\algoName, TATS, etc.) in Section~\ref{sec:method}. Finally, the experimental evaluation of \algoName is presented in Section~\ref{sec:result}.


\section{Background and Related Work}\label{sec:related}

\paragraph{\textbf{Extreme Multi-Label (XML)}} 
 Several strategies have been proposed to address the challenges of XML \cite{DBLP:journals/corr/abs-2011-11197}. 
The first category, \textit{Tree-based} methods, tackles XML by recursively partitioning the label set, and training classifiers to map each document to the appropriate sub-partition \cite{DBLP:conf/kdd/PrabhuV14, DBLP:conf/kdd/JainPV16, DBLP:conf/wsdm/PrabhuKGDHAV18}.
The second family, \textit{Embedding} methods, aims at learning an embedding of the label, that is to say a latent low dimensional vector space.
This embedding is then used to reduce the XML problem to a more manageable classification problem, that can be solved by using e.g. nearest neighbors algorithms \cite{DBLP:conf/nips/BhatiaJKVJ15, DBLP:conf/kdd/Tagami17, DBLP:conf/aaai/0001WNK0R19}. 
More recently, many works have highlighted the potential of deep learning architectures for the XML paradigm. 
One of the seminal approaches in this context is XML-CNN \cite{DBLP:conf/sigir/LiuCWY17}, where the authors apply a convolutional neural network (CNN) to learn the text representation and map it to the label space.
Some of the notable following works include AttentionXML \cite{DBLP:conf/nips/YouZWDMZ19}, which introduced an attention based architecture and leveraged shallow probabilistic label trees (PLT) to achieve better performance in XML, as well as
X-Transformer \cite{DBLP:conf/kdd/ChangYZYD20}, which first used Transformer models to tackle the XML task, and whose results were later improved by \citep{zhang2021fast} through the use of recursive fine-tuning.
Transformers were found to be a promising architecture for XML, and  \citep{kementchedjhieva2023exploration} showed that models using a Seq2Seq approach tend to perform better for such tasks.

In parallel, there has been a significant interest into the use of the structure of labels, such as taxonomies, to enhance XML methods.
For instance, \citep{gargiulo2019exploit} proposed HDNN, which incorporates the tree of labels directly into the architecture of the neural network, and which was improved in \cite{DBLP:journals/asc/GargiuloSCP19} by using a CNN.
More recent works have combined several of the aforementioned ideas, to achieve even higher performance.
Notably, MATCH~\cite{DBLP:conf/www/ZhangSDW021} used a transformer architecture while leveraging the label structure through the use of regularization, by enforcing each label to be similar to its parents, while Caled et al.~\cite{DBLP:journals/jodl/CaledSMW22} introduced a recurrent neural network, whose prediction are combined across levels of the taxonomy to improve predictions. 
Arguably the model closest to us is Hector \cite{ostapuk2024follow}, where the authors proposed an XMLCo algorithm that leverages the taxonomy tree by using a transformer architecture to directly predict a path on the label structure, using known labels to form the path prefix. 
Their approach allowed to leverage the full Seq2Seq potential of transformers and resulted in state-of-art performance for the XMLCo problem.
However, while we also use a modified transformer architecture to predict paths on a label structure, compared to their work, we extend the framework to a much more general structure, Weak-SemiLattice, that better captures the subtleties of taxonomies, and we use the resulting structure with a completely different approach based on Parallel Feature Sharing and Taxonomy Aware Tasks.
Our experiments show that our method, \algoName, outperforms existing XMLCo algorithms that are associated with information-rich taxonomies.

\paragraph*{\textbf{Multi-task Learning}} 
The paradigm of Multi-task learning (MTL) has encountered significant success since the seminal work of \cite{caruana1997multitask}.
Many methods have been developed around leveraging multiple dependent smaller tasks to improve the performance of the resulting larger model \cite{zhang2021survey,zhang2018overview}, and examples of successful MTL applications uses include 
immediacy prediction \cite{chu2015multi},
cancer detection \cite{cuccu2022typhon,cuccu2020hydra},
web search ranking \cite{chapelle2010multi},
and NLP tasks \cite{chen2024multi} such as Sequence Tagging \cite{augenstein2017multi} or Text Generation \cite{chang2020zero}.  
Feature sharing MTL is arguably one of the most popular MTL approaches in deep learning architectures, and while \algoName approach takes inspiration from it, to the best of our knowledge, \algoName is the first to intertwine MTL into the taxonomy structure for XMLCo.
Furthermore, one of our main contributions is a new taxonomy-adapted loss (Section \ref{sec:method}), and our results (see Section \ref{sec:result}) show that it is indeed a promising methodology to leverage for taxonomy-based XMLC.

\paragraph*{\textbf{Few shot Learning and XML}}
The few-shot classification problem is part of the Meta-Learning paradigm, where models are trained to recognize new classes with only a limited number of training examples.
To achieve this, the models are generally first primed with a larger dataset, before being presented with never-seen classes \cite{song2023comprehensive}. 
Some of the most influential methods in this field are arguably MAML \cite{finn2017model}, a model agnostic approach that uses a two-step gradient optimization process, and  PROTONET \cite{snell2017prototypical}, which learns an embedding space on which classes can be more easily differentiated. 
While these methods and their improvements achieve state-of-the-art results on few-shots classification problems, they generally perform poorly on XMLC due to the large numbers of independent labels.
Thus, a few  approaches have been proposed to tackle this problem.
For instance, PfastreXML \cite{jain2016extreme} addresses the rarity of the newly labels by designing a loss that emphasizes tail labels.
From a different perspective, DECAF \cite{mittal2021decaf}  leverages the new label features to improve the learning of the tail distribution.
However, while these models  achieve reasonable performance on few-shot tasks, they generally yield sup-optimal results in large XMLC problems endowed with taxonomies, contrarily to \algoName. 
Furthermore, \algoName offers a new approach to few-shot XML with the introduction of new Taxonomy-Aware Tasks.

\section{Method}\label{sec:method}

\begin{figure}[]
	\centering
	\scalebox{0.7}{
		\begin{tikzpicture}[shorten >=1pt,->,draw=black!50, node distance=1cm]
			\tikzstyle{every pin edge}=[<-,shorten <=1pt]
			\tikzstyle{base neuron}=[circle,draw=black,fill=black!25,minimum size=17pt,inner sep=0pt]
			\tikzstyle{neuron one}=[base neuron, fill=green!50];
			\tikzstyle{neuron two}=[base neuron, fill=red!50];
			\tikzstyle{neuron three}=[base neuron, fill=blue!50];

			\node[circle,draw=black,fill=black!25,minimum size=17pt,inner sep=0pt,fill=blue!20, text width= 1.5 cm, align= center] (rootCS) at (0,0) { \small Computer Science};
			
			\node[circle,draw=black,fill=black!25,minimum size=17pt,inner sep=0pt,fill=blue!20, text width= 1.5 cm, align= center] (LB11) at (-2.5,-2.5) { \small NLP};
			\node[circle,draw=black,fill=black!25,minimum size=17pt,inner sep=0pt,fill=blue!20, text width= 1.5 cm, align= center] (LB12) at (0,-2.5) { \small Database};
			\node[circle,draw=black,fill=black!25,minimum size=17pt,inner sep=0pt,fill=blue!20, text width= 1.5 cm, align= center] (LB13) at (2.5,-2.5) { \small Machine Learning};
			
			\node[circle,draw=black,fill=black!25,minimum size=17pt,inner sep=0pt,fill=blue!20, text width= 1.5 cm, align= center] (LB21) at (-3.5,-5) { \small Vocabulary  };
			\node[circle,draw=black,fill=black!25,minimum size=17pt,inner sep=0pt,fill=blue!20, text width= 1.5 cm, align= center] (LB22) at (-1.25,-5) { \small LLMs};
			\node[circle,draw=black,fill=black!25,minimum size=17pt,inner sep=0pt,fill=blue!20, text width= 1.5 cm, align= center] (LB23) at (1.25,-5) { \small Reinforcement Learning};
			\node[circle,draw=black,fill=black!25,minimum size=17pt,inner sep=0pt,fill=blue!20, text width= 1.5 cm, align= center] (LB24) at (3.5,-5) { \small Unsupervised Learning};

			%
			\draw[line width=0.6mm,draw=black] (rootCS)--(LB11);
			\draw[line width=0.6mm,draw=black] (rootCS)--(LB12);
			\draw[line width=0.6mm,draw=black] (rootCS)--(LB13);
			
			\draw[line width=0.6mm,draw=black] (LB11)--(LB21);
			\draw[line width=0.6mm,draw=black] (LB11)--(LB22);
			
			\draw[line width=0.6mm,draw=black] (LB13)--(LB22);
			\draw[line width=0.6mm,draw=black] (LB13)--(LB23);
			\draw[line width=0.6mm,draw=black] (LB13)--(LB24);
			%
			%

		\end{tikzpicture}
	}
	
	\caption{Example of a toy taxonomy of scientific labels.  An arrow from $\ell_1$ to $\ell_2$ represents $\ell_1 \leq \ell_2$.  This taxonomy can be represented with a Weak-Semilattice, and not with a tree,  as the label ``LLMs'' has multiple parents.}
	\label{fig:taxonomy_semillatice}
\end{figure}
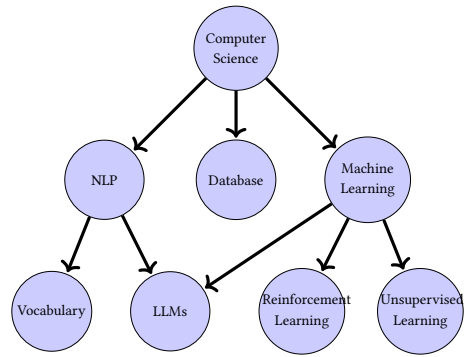

This section introduces the main contribution of this paper, i.e. Taxonomy-Aware Tasks and \algoName.

\subsection{Weak-Semilattice and Taxonomy-Aware Tasks }\label{subsec:semilattice}

Throughout this paper, we assume that the XML problem comes with a taxonomy $T$, that can be modeled as a Weak-Semilattice. We begin by briefly recalling the necessary definitions.

\begin{definition}[Partially Ordered Set]\label{def:poset}
	Let $T$ be a set endowed with a binary relation $\leq$. Then $T$ is a partially ordered set, or Poset,  if and only if $\leq$ is transitive, reflexive and antisymmetric.
\end{definition}

 In the context of a taxonomy T, the binary relation generally represents a hierarchical relationship.
  In this case, the fact that a label $\ell_1 \in T$ is more general than a label $\ell_2 \in T$ will be denoted as $\ell_1 \le \ell_2.$ 
  For example, if $T$ represents a scientific label  taxonomy (see Figure \ref{fig:taxonomy_semillatice} for a toy example), $\ell_1 =$ NLP  and $\ell_2 =$LLMs, then $\ell_1 \leq \ell_2.$ 
  It is easy to see that such binary relations would satisfy the transitivity, reflexivity and antisymmetric properties of a partial order.
\begin{definition}[Weak-Semilattice]\label{def:weak-semilattice}
	A partially ordered set $(T,\leq)$ is called a Weak-Semilattice if 
	\begin{equation*}
		\forall T' \subset T, \, \exists \ell \in T \text{ such that } \forall \ell' \in T', \quad \ell \le \ell'
	\end{equation*}
The set of elements that are smaller than $T'$, called the lower set of $T',$ is noted $\low(T')$.
\end{definition}

As a consequence, any two elements of $T$ have at least one common lower bound in $T$. 
For a hierarchical taxonomy $T,$ this lower bound can be seen as a label that is more general than any label in $T'$. 
In the toy taxonomy of scientific labels depicted in Figure \ref{fig:taxonomy_semillatice}, the greatest lower bound of Vocabulary and Machine Learning would be Computer Science. 

\paragraph{\textbf{Relation with other structures}}
Importantly, trees are a special case of Weak-Semilattice.
 Indeed, by defining  $\ell_1 \leq \ell_2$ if and only if $\ell_1$ is an ancestor of $\ell_2$,  it is easy  to see that a tree with the binary relation $\le$ is a Poset.
 Moreover, the lower set of any set of node of the tree contains the root of the tree, so the condition of Definition \ref{def:weak-semilattice} is satisfied.
The inverse is not true, as illustrated by  Fig.~\ref{fig:taxonomy_semillatice} since some elements of the Weak-Semilattice (in this case the label ``LLMs'') can have multiple parents.
We argue that this setting is quite natural in document labelling, as fine-grained labels can inherit from many general labels.
For instance, multidisciplinary scientific topics generally can stem from multiple research fields, and the same phenomenon can be observed in many taxonomies \cite{chalkidis2019large,chalkidis-etal-2021-multieurlex, DBLP:journals/cbm/RomeroNFRV23} - in particular in the MeSH taxonomy, see Section \ref{sec:result}. 
 Importantly, Hierarchical Taxonomy-based methods such as Hector \cite{ostapuk2024follow} rely on the tree structure to achieve their label prediction. 
 This is not the case of our algorithm, \algoName, that is designed to accommodate any Weak-Semilattice structure.  
 Similarly, it is easy to see that Semilattices are a special case of Weak-Semilattice, as they require the existence of an unique infimum \cite{chajda2007semilattice}.
 Finally, the relation between Weak-Semilattices and Posets is summarized in the following Lemma :
  \begin{lemma}\label{lemma:poset_and_weak_semilattice}
  	Let $(T,\leq)$ be a Poset. Then  $(T,\leq)$ has a Condorcet winner if and only if $(T,\leq)$ is a Weak-Semilattice
 \end{lemma}
 \begin{proof}
 	Let $(T,\leq)$ be a Poset. If $(T,\leq)$ has a Condorcet winner $c,$ then $\forall T' \subset T,$ $\forall \ell \in T',$ $c \leq \ell.$ Hence $(T,\leq)$ is a Weak-Semilattice.
 	Conversely, if $(T,\leq)$ is a Weak-Semilattice, let $T'=T,$ then by Definition \ref{def:weak-semilattice} $\low(T)\neq \emptyset.$ Let $c \in \low(T)$, then by definition of $\low(T)$ $c$ is a Condorcet winner. 
 	\end{proof}

\paragraph{\textbf{Children and Width of a Weak-Semilattice}} We also need to define the children and the width for a Weak-Semilattice, as they will be key to \algoName training.
We define them as extensions of their respective notions in a tree.
\begin{definition}[Children in a Weak-Semilattice]
	Let $(T,\leq)$ be a Weak-Semilattice and $\ell_1 \in  T.$  Let $\ell_2 \in T$ such that $\ell_1 \le \ell_2$ and $\ell_1\neq\ell_2$. $\ell_2$ is called a child of $\ell_1$ (denoted $\ell_1 \prec \ell_2$) if and only if $\forall \ell \in T,$ if $\ell_1 \le \ell \le \ell_2$ then $\ell_1=\ell$ or $\ell_2=\ell.$  
\end{definition}

\begin{definition}[Width of a Weak-Semilattice]\label{def:width}
	 Let $(T,\leq)$ be a Weak-Semilattice.  The width of the Weak-Semilattice $w_T$ is then defined as 	 
	 $$w_T = \max_{\ell \in T} \# \left\lbrace \ell' \in T, \quad \ell \prec \ell' \right\rbrace $$  
\end{definition}
In other words, the width of $T$ is the maximal number of children of any element of $T.$ 
This notion directly relates to the difficulty of predicting a path on $T,$ as it translates into the maximum possible number of labels to choose from extending the path.
%

\paragraph{\textbf{Taxonomy-Aware Tasks}.}\label{subsec:tats}
At the heart of our approach is the decomposition of the XMLCo problem into Taxonomy-Aware Tasks (TATs).
Such a decomposition allows to better leverage the Parallel Feature Sharing framework of the Multi-Task paradigm  by adjusting the tasks-specific components of \algoName.
Furthermore, as each resulting task contains significantly fewer labels, and thus a better document / label ratio, this approach alleviates the data scarcity problem.
However, to achieve these advantages, the splitting of the taxonomy must be coherent with the Weak-Semilattice structure, and not remove valuable information regarding the relation between labels. 
In particular, such decomposition should avoid breaking paths between multiple sub-tasks, as the prediction of paths is at the heart of our method, \algoName. 
This is ensured by the following conditions. 

\begin{definition}[Taxonomy Aware Tasks]\label{def:tats}
Let $(T,\leq)$ be a Weak-Semilattice. Then the collection of sets $(T_1, \ldots, T_N)$ are Taxonomy Aware Tasks if and only if: 	
	\begin{enumerate}
	\item $\forall 1\le i \le N,$  $T_i \subset T$ and $(T_i,\le)$ is a Weak-Semilattice,
	\item $\forall 1\le i,j \le N,$  if $T_i \subset T_j$ then $T_i=T_j,$
	\item $\forall 1\le i \le N,$  $\forall \ell \in T_i,$ $\forall \ell' \in T,$ if  $\ell \le \ell'$ then  $\ell' \in T_i.$ 
	\item $\forall \ell \in T \setminus \low(T),$  $\exists  1\le i \le N$ such that $\ell \in T_i.$ 
	
\end{enumerate}
\end{definition}
The first and third conditions enforce the preservation of paths: if one label $\ell$ is present in $T_i$ then all the paths that originate from  $\ell$ are also in $T_i$.
This is key to \algoName as it predicts labels by forecasting paths in the Weak-Semilattice structure.
The second condition prevents repeat tasks, while the fourth ensures that all the labels, with the exception of the Condorcet winner, are part of at least one task.

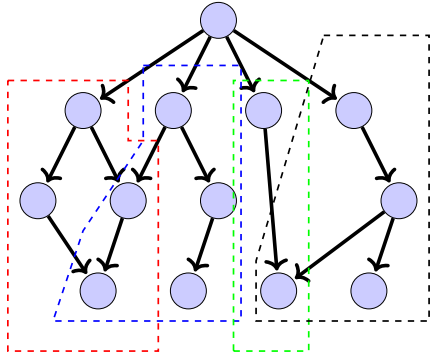
\begin{figure}[]
	\centering
	\scalebox{0.8}{
		\begin{tikzpicture}[shorten >=1pt,draw=black!50, node distance=1cm]
			\tikzstyle{base neuron}=[circle,draw=black,fill=black!25,minimum size=17pt,inner sep=0pt]
			\tikzstyle{neuron one}=[base neuron, fill=green!50];
			\tikzstyle{neuron two}=[base neuron, fill=red!50];
			\tikzstyle{neuron three}=[base neuron, fill=blue!50];

			\node[circle,draw=black,fill=black!25,minimum size=17pt,inner sep=0pt,fill=blue!20, text width= .5 cm, align= center] (rootCS) at (0,0) { };
			
			\node[circle,draw=black,fill=black!25,minimum size=17pt,inner sep=0pt,fill=blue!20, text width= .5 cm, align= center] (L11) at (-2.25,-1.5) { };
			\node[circle,draw=black,fill=black!25,minimum size=17pt,inner sep=0pt,fill=blue!20, text width= .5 cm, align= center] (L12) at (-.75,-1.5) { };
			\node[circle,draw=black,fill=black!25,minimum size=17pt,inner sep=0pt,fill=blue!20, text width= .5 cm, align= center] (L13) at (.75,-1.5) { };
			\node[circle,draw=black,fill=black!25,minimum size=17pt,inner sep=0pt,fill=blue!20, text width= .5 cm, align= center] (L14) at (2.25,-1.5) { };
			
			\node[circle,draw=black,fill=black!25,minimum size=17pt,inner sep=0pt,fill=blue!20, text width= .5 cm, align= center] (L21) at (-3,-3) { };
			\node[circle,draw=black,fill=black!25,minimum size=17pt,inner sep=0pt,fill=blue!20, text width= .5 cm, align= center] (L22) at (-1.5,-3) { };
			\node[circle,draw=black,fill=black!25,minimum size=17pt,inner sep=0pt,fill=blue!20, text width= .5 cm, align= center] (L23) at (0,-3) { };
			\node[circle,draw=black,fill=black!25,minimum size=17pt,inner sep=0pt,fill=blue!20, text width= .5 cm, align= center] (L24) at (3,-3) { };
			
			\node[circle,draw=black,fill=black!25,minimum size=17pt,inner sep=0pt,fill=blue!20, text width= .5 cm, align= center] (L31) at (-2,-4.5) { };
			\node[circle,draw=black,fill=black!25,minimum size=17pt,inner sep=0pt,fill=blue!20, text width= .5 cm, align= center] (L32) at (-0.5,-4.5) { };
			\node[circle,draw=black,fill=black!25,minimum size=17pt,inner sep=0pt,fill=blue!20, text width= .5 cm, align= center] (L33) at (1,-4.5) { };
			\node[circle,draw=black,fill=black!25,minimum size=17pt,inner sep=0pt,fill=blue!20, text width= .5 cm, align= center] (L34) at (2.5,-4.5) { };

			%
			\draw[line width=0.6mm,draw=black,->] (rootCS)--(L11);
			\draw[line width=0.6mm,draw=black,->] (rootCS)--(L12);
			\draw[line width=0.6mm,draw=black,->] (rootCS)--(L13);
			\draw[line width=0.6mm,draw=black,->] (rootCS)--(L14);

			\draw[line width=0.6mm,draw=black,->] (L11)--(L21);
			\draw[line width=0.6mm,draw=black,->] (L11)--(L22);
			\draw[line width=0.6mm,draw=black,->] (L12)--(L22);
			\draw[line width=0.6mm,draw=black,->] (L12)--(L23);
			\draw[line width=0.6mm,draw=black,->] (L14)--(L24);
			
			\draw[line width=0.6mm,draw=black,->] (L21)--(L31);
			\draw[line width=0.6mm,draw=black,->] (L22)--(L31);
			\draw[line width=0.6mm,draw=black,->] (L23)--(L32);
			\draw[line width=0.6mm,draw=black,->] (L24)--(L34);
			\draw[line width=0.6mm,draw=black,->] (L24)--(L33);
			\draw[line width=0.6mm,draw=black,->] (L13)--(L33);

			\def\delt{.5}

			\draw[draw, thick, red,dashed] (-3 -\delt, -1.5 + \delt) --  (-.75 - 1.5*\delt, -1.5 + \delt) -- (-.75 -1.5*\delt, -1.5 -\delt) -- (-.75 -0.5*\delt, -1.5 -\delt) --   (-0.75 - 0.5*\delt,-4.5-2*\delt) -- (-3 - \delt,-4.5-2*\delt) --cycle ;
			
			\draw[draw, thick, blue,dashed]   (-.75 - \delt, -1.5 + 1.5*\delt) -- (.75-0.75*\delt, -1.5 +1.5*\delt) --
			 (.75  - 0.75 *\delt, -4.5 -\delt) -- (-2.25 - \delt , -4.5 -\delt) -- (-2.25   , -3 -\delt)  --  (-.75 - \delt, -1.5 - \delt) -- cycle ;
			 
			 \draw[draw, thick, green,dashed]   (.75 - \delt, -1.5 +\delt) -- (1+\delt, -1.5 +\delt) --
			  (1+\delt, -4.5 - 2*\delt) -- (0.75-\delt, -4.5 - 2*\delt) -- cycle ;
			  \draw[draw, thick, black,dashed]   (2.25 - \delt, -1.5 +2.5*\delt) -- (3+\delt, -1.5 +2.5*\delt) --
			  (3+\delt, -4.5 - \delt) -- (1-0.75*\delt, -4.5 - \delt)  --   (1-0.75*\delt, -4.5 + \delt) -- cycle ;

			

			%
			%

		\end{tikzpicture}
	}
	
	\caption{Example of a Weak Semilattice taxonomy with a Taxonomy Aware Tasks decomposition, represented by the different polygons. }
	\label{fig:tats}
\end{figure}

Figure \ref{fig:tats} shows an example of a Weak-Semilattice and a TATs decomposition.
Note that the combination of the conditions in Definition \ref{def:tats} makes TATs decomposition unique.
%
%
Interestingly, in the case where $T$ is a tree, Definition \ref{def:tats} implies that the TATs decomposition will be made of all the subtrees of $T$ that have node of depth one as root.  
While in the tree setting the TAT are disjoints (i.e. they share no common labels), this is generally not true for Weak-Semilattice as illustrated by Figure \ref{fig:tats}.

\subsection{\algoName}\label{subsec:algo}

Similarly to Hector \cite{ostapuk2024follow}, \algoName uses a modified transformer to predict  paths on the taxonomy, that are then combined to obtain the algorithm predictions. 
We begin by recalling the setup of path prediction, before discussing the specificities entailed by the Weak-Semilattice structure and our TATs-based approach. 

\medskip

\noindent \textbf{Path Prediction.}
While the labels of a document are generally encoded as a set, they can also be represented as a collection of paths in the taxonomy, from the most general to the more specific \cite{ostapuk2024follow}.
In the context of a Weak-Semilattice, a path is defined as a sequence of labels and their children, i.e. $\ell_1 \prec \ell_2 \prec \ldots \prec \ell_K,$ where $\ell_1$ is the Condorcet winner, or ``root'', of the taxonomy (see Lemma \ref{lemma:poset_and_weak_semilattice}).
Such a path can be seen as a sequence of increasingly specific labels  that characterize the documents. 

Predicting paths has multiple advantages over predicting a set of labels.
First, these paths yield a natural sequential structure, contrarily to sets, thus allowing to fully leverage the transformer architecture, which has been shown to achieve state-of-the-art performance on XML problems \cite{DBLP:conf/nips/YouZWDMZ19, zhang2021fast, ostapuk2024follow}. 
Second, this approach naturally embeds the taxonomy structure into the prediction, as at each step, only the children of a label will be considered to be added to the path.
This both encodes the relation between labels (a key ingredient of successful XML methods, see e.g. \cite{DBLP:conf/www/ZhangSDW021}) and also alleviates the label scale problem by strongly reducing the number of candidate at each step.

A key difference with previous works is that \algoName predicts paths on TATs, which are sub-taxonomies  of $T$.
Depending on the path prefix, one or more TATs may be relevant to the predictions at hand. 
Thus \algoName predicts paths in parallel across all the relevant TATs.
Moreover, even when only one TAT is compatible with a given label, multiple paths may lead to this label.
The combination of predictions across multiple paths and TATs is discussed in the Label Prediction paragraph below.
%

\medskip

\noindent {\textbf{Model Architecture.}}
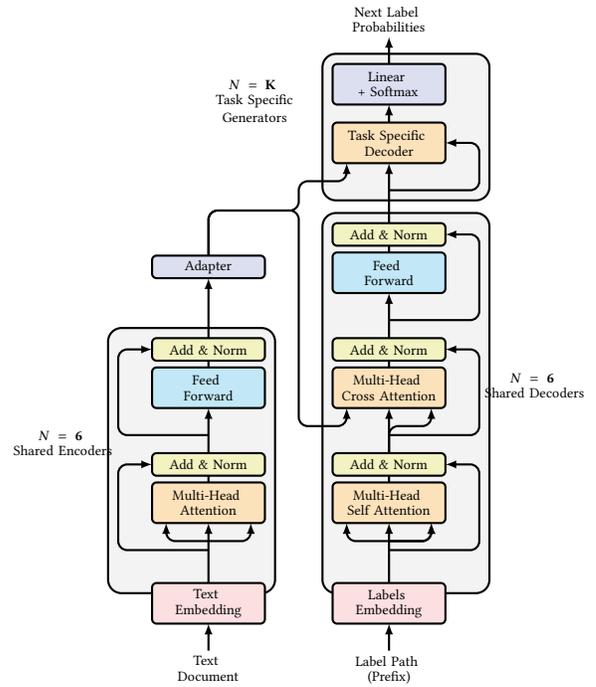
\begin{figure}[]
	\centering
	\scalebox{0.6}{
	\begin{tikzpicture}
\definecolor{emb_color}{RGB}{252,224,225}
\definecolor{multi_head_attention_color}{RGB}{252,226,187}
\definecolor{add_norm_color}{RGB}{242,243,193}
\definecolor{ff_color}{RGB}{194,232,247}
\definecolor{softmax_color}{RGB}{203,231,207}
\definecolor{linear_color}{RGB}{220,223,240}
\definecolor{gray_bbox_color}{RGB}{243,243,244}
\draw[fill=gray_bbox_color, line width=0.046875cm, rounded corners=0.300000cm] (-0.975000, 5.655000) -- (2.725000, 5.655000) -- (2.725000, -0.225000) -- (-0.975000, -0.225000) -- cycle;
\draw[fill=gray_bbox_color, line width=0.046875cm, rounded corners=0.300000cm] (3.775000, 8.205000) -- (7.475000, 8.205000) -- (7.475000, -0.225000) -- (3.775000, -0.225000) -- cycle;
\draw[fill=gray_bbox_color, line width=0.046875cm, rounded corners=0.300000cm] (3.775000, 11.735000) -- (7.475000, 11.735000) -- (7.475000, 8.485000) -- (3.775000, 8.485000) -- cycle;
\draw[line width=0.046875cm, fill=emb_color, rounded corners=0.100000cm] (0.000000, 0.000000) -- (2.500000, 0.000000) -- (2.500000, -0.900000) -- (0.000000, -0.900000) -- cycle;
\node[text width=2.500000cm, align=center] at (1.250000,-0.450000) {Text \vspace{-0.05cm} \linebreak Embedding};
\draw[line width=0.046875cm, fill=emb_color, rounded corners=0.100000cm] (4.000000, 0.000000) -- (6.500000, 0.000000) -- (6.500000, -0.900000) -- (4.000000, -0.900000) -- cycle;
\node[text width=2.500000cm, align=center] at (5.250000,-0.450000) {Labels \vspace{-0.05cm} \linebreak Embedding};
\draw[line width=0.046875cm, fill=add_norm_color, rounded corners=0.100000cm] (0.000000, 2.880000) -- (2.500000, 2.880000) -- (2.500000, 2.380000) -- (0.000000, 2.380000) -- cycle;
\node[text width=2.500000cm, align=center] at (1.250000,2.630000) {Add \& Norm};
\draw[line width=0.046875cm, fill=multi_head_attention_color, rounded corners=0.100000cm] (0.000000, 2.230000) -- (2.500000, 2.230000) -- (2.500000, 1.330000) -- (0.000000, 1.330000) -- cycle;
\node[text width=2.500000cm, align=center] at (1.250000,1.780000) {Multi-Head \vspace{-0.05cm} \linebreak Attention};
\draw[line width=0.046875cm] (1.250000, 2.230000) -- (1.250000, 2.380000);
\draw[line width=0.046875cm, fill=add_norm_color, rounded corners=0.100000cm] (4.000000, 5.430000) -- (6.500000, 5.430000) -- (6.500000, 4.930000) -- (4.000000, 4.930000) -- cycle;
\node[text width=2.500000cm, align=center] at (5.250000,5.180000) {Add \& Norm};
\draw[line width=0.046875cm, fill=multi_head_attention_color, rounded corners=0.100000cm] (4.000000, 4.780000) -- (6.500000, 4.780000) -- (6.500000, 3.880000) -- (4.000000, 3.880000) -- cycle;
\node[text width=2.500000cm, align=center] at (5.250000,4.330000) {Multi-Head \vspace{-0.05cm} \linebreak Cross Attention};
\draw[line width=0.046875cm] (5.250000, 4.780000) -- (5.250000, 4.930000);
\draw[line width=0.046875cm, fill=add_norm_color, rounded corners=0.100000cm] (4.000000, 2.880000) -- (6.500000, 2.880000) -- (6.500000, 2.380000) -- (4.000000, 2.380000) -- cycle;
\node[text width=2.500000cm, align=center] at (5.250000,2.630000) {Add \& Norm};
\draw[line width=0.046875cm, fill=multi_head_attention_color, rounded corners=0.100000cm] (4.000000, 2.230000) -- (6.500000, 2.230000) -- (6.500000, 1.330000) -- (4.000000, 1.330000) -- cycle;
\node[text width=2.500000cm, align=center] at (5.250000,1.780000) {Multi-Head \vspace{-0.05cm} \linebreak Self Attention};
\draw[line width=0.046875cm] (5.250000, 2.230000) -- (5.250000, 2.380000);
\draw[line width=0.046875cm, fill=add_norm_color, rounded corners=0.100000cm] (0.000000, 5.430000) -- (2.500000, 5.430000) -- (2.500000, 4.930000) -- (0.000000, 4.930000) -- cycle;
\node[text width=2.500000cm, align=center] at (1.250000,5.180000) {Add \& Norm};
\draw[line width=0.046875cm, fill=ff_color, rounded corners=0.100000cm] (0.000000, 4.780000) -- (2.500000, 4.780000) -- (2.500000, 3.880000) -- (0.000000, 3.880000) -- cycle;
\node[text width=2.500000cm, align=center] at (1.250000,4.330000) {Feed \vspace{-0.05cm} \linebreak Forward};
\draw[line width=0.046875cm] (1.250000, 4.780000) -- (1.250000, 4.930000);
\draw[line width=0.046875cm, fill=add_norm_color, rounded corners=0.100000cm] (4.000000, 7.980000) -- (6.500000, 7.980000) -- (6.500000, 7.480000) -- (4.000000, 7.480000) -- cycle;
\node[text width=2.500000cm, align=center] at (5.250000,7.730000) {Add \& Norm};
\draw[line width=0.046875cm, fill=ff_color, rounded corners=0.100000cm] (4.000000, 7.330000) -- (6.500000, 7.330000) -- (6.500000, 6.430000) -- (4.000000, 6.430000) -- cycle;
\node[text width=2.500000cm, align=center] at (5.250000,6.880000) {Feed \vspace{-0.05cm} \linebreak Forward};
\draw[line width=0.046875cm] (5.250000, 7.330000) -- (5.250000, 7.480000);
\draw[line width=0.046875cm, fill=multi_head_attention_color, rounded corners=0.100000cm] (4.000000, 10.210000) -- (6.500000, 10.210000) -- (6.500000, 9.310000) -- (4.000000, 9.310000) -- cycle;
\node[text width=2.500000cm, align=center] at (5.250000,9.760000) {Task Specific \vspace{-0.05cm} \linebreak Decoder};
\draw[line width=0.046875cm, fill=linear_color, rounded corners=0.100000cm] (4.000000, 11.510000) -- (6.500000, 11.510000) -- (6.500000, 10.610000) -- (4.000000, 10.610000) -- cycle;
\node[text width=2.500000cm, align=center] at (5.250000,11.060000) {Linear \vspace{-0.05cm} \linebreak + Softmax};
\draw[line width=0.046875cm, fill=linear_color, rounded corners=0.100000cm] (0.000000, 7.260000) -- (2.500000, 7.260000) -- (2.500000, 6.760000) -- (0.000000, 6.760000) -- cycle;
\node[text width=2.500000cm, align=center] at (1.250000,7.010000) {Adapter};
\draw[line width=0.046875cm, -latex] (1.250000, 2.880000) -- (1.250000, 3.880000);
\draw[line width=0.046875cm, -latex] (5.250000, 5.430000) -- (5.250000, 6.430000);
\draw[line width=0.046875cm, -latex] (5.250000, 7.980000) -- (5.250000, 9.310000);
\draw[line width=0.046875cm, -latex] (5.250000, 10.210000) -- (5.250000, 10.610000);
\draw[line width=0.046875cm, -latex] (1.250000, 0.000000) -- (1.250000, 1.330000);
\draw[line width=0.046875cm, -latex] (5.250000, 0.000000) -- (5.250000, 1.330000);
\draw[line width=0.046875cm, -latex] (5.250000, 2.880000) -- (5.250000, 3.880000);
\draw[line width=0.046875cm, -latex] (1.250000, 5.430000) -- (1.250000, 6.760000);
\draw[-latex, line width=0.046875cm, rounded corners=0.200000cm] (1.250000, 3.280000) -- (-0.750000, 3.280000) -- (-0.750000, 5.180000) -- (0.000000, 5.180000);
\draw[-latex, line width=0.046875cm, rounded corners=0.200000cm] (1.250000, 0.730000) -- (-0.750000, 0.730000) -- (-0.750000, 2.630000) -- (0.000000, 2.630000);
\draw[-latex, line width=0.046875cm, rounded corners=0.200000cm] (5.250000, 0.730000) -- (7.250000, 0.730000) -- (7.250000, 2.630000) -- (6.500000, 2.630000);
\draw[-latex, line width=0.046875cm, rounded corners=0.200000cm] (5.250000, 3.280000) -- (7.250000, 3.280000) -- (7.250000, 5.180000) -- (6.500000, 5.180000);
\draw[-latex, line width=0.046875cm, rounded corners=0.200000cm] (5.250000, 5.830000) -- (7.250000, 5.830000) -- (7.250000, 7.730000) -- (6.500000, 7.730000);
\draw[-latex, line width=0.046875cm, rounded corners=0.200000cm] (5.250000, 8.710000) -- (7.250000, 8.710000) -- (7.250000, 9.760000) -- (6.500000, 9.760000);
\draw[-latex, line width=0.046875cm, rounded corners=0.200000cm] (1.250000, 0.930000) -- (0.312500, 0.930000) -- (0.312500, 1.330000);
\draw[-latex, line width=0.046875cm, rounded corners=0.200000cm] (1.250000, 0.930000) -- (2.187500, 0.930000) -- (2.187500, 1.330000);
\draw[-latex, line width=0.046875cm, rounded corners=0.200000cm] (5.250000, 0.930000) -- (4.312500, 0.930000) -- (4.312500, 1.330000);
\draw[-latex, line width=0.046875cm, rounded corners=0.200000cm] (5.250000, 0.930000) -- (6.187500, 0.930000) -- (6.187500, 1.330000);
\draw[-latex, line width=0.046875cm, rounded corners=0.200000cm] (1.250000, 7.260000) -- (1.250000, 8.260000) -- (3.250000, 8.260000) -- (3.250000, 3.480000) -- (4.312500, 3.480000) -- (4.312500, 3.880000);
\draw[-latex, line width=0.046875cm, rounded corners=0.200000cm] (1.250000, 7.260000) -- (1.250000, 8.260000) -- (3.250000, 8.260000) -- (3.250000, 8.910000) -- (4.312500, 8.910000) -- (4.312500, 9.310000);
\draw[-latex, line width=0.046875cm, rounded corners=0.200000cm] (5.250000, 0.930000) -- (6.187500, 0.930000) -- (6.187500, 1.330000);
\draw[-latex, line width=0.046875cm, rounded corners=0.200000cm] (5.250000, 2.880000) -- (5.250000, 3.480000) -- (6.187500, 3.480000) -- (6.187500, 3.880000);
\draw[line width=0.046875cm, -latex] (1.250000, -1.500000) -- (1.250000, -0.900000);
\draw[line width=0.046875cm, -latex] (5.250000, -1.500000) -- (5.250000, -0.900000);
\draw[line width=0.046875cm, -latex] (5.250000, 11.510000) -- (5.250000, 12.110000);
\node[text width=2.500000cm, anchor=north, align=center] at (1.250000,-1.500000) {Text \vspace{-0.05cm} \linebreak Document};
\node[text width=2.500000cm, anchor=north, align=center] at (5.250000,-1.500000) {Label Path \vspace{-0.05cm} \linebreak (Prefix)};
\node[text width=2.500000cm, anchor=south, align=center] at (5.250000,12.110000) {Next Label \vspace{-0.05cm} \linebreak Probabilities};
\node[text width=2.500000cm, anchor=south, align=center] at (-1.975000,2.715000) {$ N = \textbf{6}$ \vspace{-0.05cm} \linebreak Shared Encoders};
\node[text width=2.500000cm, anchor=south, align=center] at (8.475000,3.990000) {$ N = \textbf{6}$ \vspace{-0.05cm} \linebreak Shared Decoders};
\node[text width=2.500000cm, anchor=south, align=center] at (2.275000,10.110000) {$ N = \textbf{K}$ \vspace{-0.05cm} \linebreak Task Specific Generators};
\end{tikzpicture}
}
	\caption{\algoName's architecture. The model is made of 6 Encoders and 6 Decoders with weights shared across tasks, as well as one Task Specific Generator per task, whose weights are task specific.}
	\label{fig:architecture}
	\vspace{-0.1in}
\end{figure}
Figure \ref{fig:architecture} details \algoName's architecture, which is based on Transformers \cite{DBLP:conf/nips/VaswaniSPUJGKP17}. 
Similar to HECTOR \cite{ostapuk2024follow}, \algoName uses both Decoder and Encoder blocks, whose weights are shared across all tasks.
However, and in contrast to previous XML Transformer-based architectures, \algoName also introduces a new type of block, the Task-Specific Generator.

When predicting the next label of a path, \algoName receives as input the text of the document, the current path as well as the current TAT of interest.
The document is first encoded using a 300 dimensional embedding, which is then fed to a stack of six Encoder blocks.
Each Encoder block contains a multi-head attention layer, with 12 Heads, that captures the contextual information of the different tokens of the document, followed by a fully connected feed-forward layer with residual connection.
At the end of the six Encoders block, the resulting 300 dimensional encoding is then projected into a 600 dimensional encoding using a fully connected adapter layer. 
In parallel, the label path is encoded using a 600 dimensional embedding, and then processed through a six Decoder blocks.
Each Decoder block is made of a multi-head self attention layer, with 12 Heads, that considers the prefix of the path to provide better prediction to the next label, followed by a cross attention layer, which captures the dependency between the document encoding and the path encoding, followed in turn by a fully connected feed-forward layer with residual connection.

Finally, at the end of the six Decoder blocks, the resulting 600 dimensional encoding is then send to the Task-Specific Generator block corresponding to the TAT of interest.
This block contains a task-specific decoder block, where labels in the label path that do not belong to the TAT are masked.
This allows to focus on TAT-relevant label encoding.  
This layer is followed by a fully connected feed forward layer that maps to the space of all possible labels that belong to the TAT.

Before training, the weights of \algoName's layers are initialized randomly, with the exception of the text embedding where we used pre-trained GloVe embeddings \cite{pennington2014glove}.

\medskip

\noindent {\textbf{Training \algoName}}
In the following we assume that the set of label of each document $\mathcal{D}$ is complete, that is to say $\mathcal{D}$ contains the labels necessary to form a path to each label of $\mathcal{D}$.
This is in line with the usual label completion assumption: if a document has a specific label $\ell$, it should also possess broader labels that contain the sub-label $\ell$ \cite{DBLP:journals/cbm/RomeroNFRV23, ostapuk2024follow}. 
If this assumption is not satisfied, the missing labels are added during the preprocessing of the dataset, in line with Hierarchical Label Set Expansion proposed by \cite{DBLP:journals/asc/GargiuloSCP19}. 
In the case where a label $\ell \in \mathcal{D}$ has no valid path in $\mathcal{D}$ but multiple possible paths exists in the full taxonomy $T,$ we add the minimum number of labels possible to obtain at least one valid path.
Ties between two possible paths requiring the same number of additional labels are broken at random.  

Then, each document's collection of labels $\mathcal{D}$ is transformed into a collection of paths $\mathcal{P}$. 
Each path $p\in \mathcal{P}$ is subsequently associated to one or more relevant tasks. 
A TAT $T_i$ is said to be relevant to a path $p$ if there exists a label $\ell\in p$ such that $\ell \in T_i.$ 
In other words, a task is deemed relevant to a path if it has at least one label of this path.

\noindent \textbf{Loss Function.} To train \algoName we introduce a new TAT-dependent loss function $\mathcal{L}$.
This loss is based on cross-entropy with both label smoothing with a value $\varepsilon = 0.01$ \cite{DBLP:conf/cvpr/SzegedyVISW16}, with a confidence decreased proportionally to the width of the relevant task. 
Formally, for a given document  $d$, a path prefix $p$, next label $\ell,$ relevant task $T_i$ and predicted probability distribution over the next label  $\hat{\ell}$, the loss is defined as:
\begin{align*}\label{eq:loss tamlec}
\mathcal{L}(\hat{\ell}, \ell , T_i) &=     \left(1- \varepsilon\frac{ w_{T_i}}{1+w_{T_i}}\right) \log\left( P(\hat{\ell} = \ell) \right)   \\
& \quad +   \sum_{\ell' \in T_i, \ell' \neq \ell}    \varepsilon\frac{w_{T_i} }{1+w_{T_i}} \log\left( 1- P(\hat{\ell} = \ell') \right)  
\end{align*}
 It is important to note that $\mathcal{L}$ only considers labels in the task $T_i$ and ignores other labels.
Compared to the original label smoothing, this loss applies a dynamic smoothing, where high confidence prediction are more favored for tasks with low width, compared to tasks with large width. 
This is in line with the idea that the formers can be seen as easier tasks, see  Definition \ref{def:width} and the following remark. 
 %

\medskip

\noindent {\textbf{Label Prediction}.}
At inference time, \algoName receives a document and an incomplete set of labels $\mathcal{D}.$
Similarly to the training phase, this set of label is transformed into a collection of paths $\mathcal{P}$ with their corresponding relevant TATs.
For each path and TAT, \algoName generate multiple path extensions using beam search, a commonly-used algorithm for decoding structured predictors, similarly to \cite{ostapuk2024follow}. 
Beam search maintains a list of the most promising candidate paths, together with their combined predicted probabilities, and iteratively updates it by feeding the paths into \algoName, until no further path can be obtained as the most specific labels have been reached.
At the end of the process, each label $\ell$ of the TAT is attributed a score equals to the sum of all the probabilities of the Beam search's predicted paths that had $\ell$ as a leaf.
Scores are then summed across all paths and relevant TATs to obtain the final score of each label, which is then used to compute the final label ranking.

\section{Experimental Evaluation}
\label{sec:result}

In this section we  empirically evaluate \algoName on both Label Completion (Section \ref{subsec:xp_XMLCO}) and Few-Shots XML tasks (Section \ref{subsec:xp_FewShots}).
\subsection{Experimental Setting}
\label{subsec:result_setting}

%
\begin{table}[!t]
	
	\centering
	\begin{tabular}{|l|lll|}
		\toprule
		& \textbf{MAG-CS} & \textbf{PubMed} & \textbf{EURLex} \\ \midrule
		N Docs &  140994 &  331720 & 172120 \\
		N Labels & 2641 & 5911 & 4492 \\
		Avg Labels per Doc & 4.4 & 18.5 & 10.4 \\
		Taxonomy Width & 145 & 21 & 145 \\\midrule
		N TATS &  24 & 8  & 145 \\
		Avg TAT Width & 10.5 & 15.5 & 3\\
		Med. Doc per TAT &   2443.33 & 34563 & 520\\
 \bottomrule
	\end{tabular}
	\caption{Important Dataset  and TATs decomposition statistics.  The three datasets exhibits very different  TATs decomposition profile, in term of number of TATs, width and number of documents per TAT. }
	\label{tab:dataset_stat}
\end{table}

\noindent\textbf{Datasets.} 
We perform our experiments on  three  commonly used XML datasets that are endowed with rich taxonomies: MAG-CS, PubMed and EURLex. 
All the datasets are completed to abide by the hierarchical taxonomy using the process described in Section \ref{subsec:algo}. 
The first dataset, MAG-CS, the Microsoft Academic Graph (MAG) Computer Science (CS)  is a subset of the MAG dataset  \cite{DBLP:journals/qss/WangSHWDK20} focused on CS, which contains papers published between 1990 and 2020 at top CS conferences \cite{DBLP:conf/www/ZhangSDW021}.
The root of the taxonomy is the label ``Computer Science'', while the most general concepts include ``Machine Learning'',  ``Natural Language Processing'' and ``World Wide Web''.
The second dataset, PubMed, was published by \cite{DBLP:conf/www/ZhangSDW021} and contains scientific papers from top medical journal published between 2010 and 2020, endowed with labels from the Medical Subject Headings (MeSH) taxonomy.
The most general concepts of the taxonomy include ``Anatomy'',  ``Psychiatry and Psychology'' as well as ``Diseases''. Importantly, the MeSH taxonomy is particularly representative of a non-tree taxonomy, as many labels inherits from multiple parents. For instance,  ``Digestive System Neoplasms'' can be reached through two distinct paths that include either ``Digestive System Diseases'' or ``Neoplasms''.
Finally, the last dataset, EURLex \cite{DBLP:conf/acl/ChalkidisFMA19}, includes EU legislative documents equipped with labels from the European Vocabulary (EuroVoc) taxonomy, which contains more than 100 high level concepts, resulting in a very large TATs decomposition compared to the other datasets.
Table~\ref{tab:dataset_stat} summarizes the key dataset characteristics, as well the description of their TATs decomposition.
Interestingly, these three datasets' TATs decompositions are very different: for instance, EURLex TATs have an average width of 3, compared to more than 10 for the two other dataset.
Similarly, PubMed TATs decompositions contain only 8 TATs, compared to 145 for EURLex. 
This diversity allows testing \algoName on substantially different use-cases.

\smallskip

\noindent\textbf{Baseline Models.}
Throughout our experiments, we compare the performance of \algoName with a large set of baselines, which includes both XML and few shots commonly-used methods:

\begin{itemize}
	\item \textsc{MATCH} \cite{DBLP:conf/www/ZhangSDW021}, an XMLC algorithm which uses both a transformer architecture and leverages the taxonomy for model regularization;
	\item \textsc{XML-CNN} \cite{DBLP:conf/sigir/LiuCWY17}, a convolutional neural network-based XMLC method,
	\item \textsc{AttentionXML} \cite{DBLP:conf/nips/YouZWDMZ19}, an XMLC method that relies on dynamically build hierarchical clustering of the labels,
	\item \textsc{Hector} \cite{ostapuk2024follow}, a recent XMLCo algorithm which uses a modified transformer architecture and path predictions on the taxonomy. \textsc{Hector} currently represents the state of the art in XMLC and was shown to be superior to previous approaches; 
	\item \textsc{MAML-T}, a combination of the Few Shots Meta Learning method MAML \cite{finn2017model} with a transformer architecture derived from MATCH; 
\end{itemize}

All baselines are trained on our modified versions of the three datasets, using the hyperparameter values recommended by their respective authors.
As  MAML is a  general few shots algorithm that require a base model adapted to the problem, we use a modified transformer architecture derived from MATCH, i.e. a three layer encoder-only neural network with two multi-heads self attention layers.
Finally, \algoName was trained using an Adam optimizer with an initial learning rate of $5 \times10^{-5}$, a and a weight decay of $10^{-2}$ and a smoothing parameter of $0.01$. 

In addition to the different baselines, we also perform an ablation study of the XMLCo performance of \algoName. We denote by  \algoAblation a tampered down version of \algoName without the adaptive loss or the advanced weight sharing. 

We will openly publish the code of our model as well as our full evaluation setup once this paper is accepted. 

\smallskip

\begin{table}[t]
	\vspace{-0.1in}
	
	\begin{tabular}{l c| ccc | ccc }
		\toprule
		\multirow{2}{*}{} & \multirow{2}{*}{\textbf{Algorithm}} & \multicolumn{3}{c}{\textbf{Precision}} & \multicolumn{3}{c}{\textbf{NDCG}} \\
		\cline{3-8}
		& & @1 & @2 & @3    &  @2 & @3 & @4     \\
		\midrule
		\parbox[t]{2mm}{\multirow{8}{*}{\rotatebox[origin=c]{90}{ MAG-CS}}} &  MATCH &  0.787  & 0.765  & 0.747  & 0.785  & 0.782  & 0.786\\
		&  XML-CNN &  0.541  & 0.578  & 0.591  & 0.607  & 0.643  & 0.674\\
		&  Att.XML &  0.810  & 0.789  & \textbf{0.778}  & 0.805  & 0.807  & 0.810\\
		&  HECTOR &  0.856  & 0.796  & 0.776  & 0.821  & {0.813}  & {0.810}\\
		&  MAML-T &  0.376  & 0.370  & 0.367  & 0.410  & 0.425  & 0.446\\
		&  $\sqrt{\text{TAMLEC}}$ &  0.830  & 0.766  & 0.730  & 0.790  & 0.768  & 0.758\\
		&  TAMLEC &  \textbf{0.870}  & \textbf{0.802}  & 0.776  & \textbf{0.827}  & \textbf{0.815}  & \textbf{0.814}\\
		\midrule
		\parbox[t]{2mm}{\multirow{8}{*}{\rotatebox[origin=c]{90}{ PUBMed}}} &  MATCH &  0.789  & 0.749  & 0.755  & 0.758  & 0.769  & 0.794\\
		&  XML-CNN &  0.773  & 0.732  & 0.741  & 0.743  & 0.760  & 0.799\\
		&  Att.XML &  0.790  & 0.757  & 0.762  & 0.764  & 0.775  & 0.798\\
		&  HECTOR &  0.866  & 0.835  & 0.836  & 0.843  & 0.849  & 0.872\\
		&  MAML-T &  0.631  & 0.575  & 0.590  & 0.590  & 0.612  & 0.683\\
		&  $\sqrt{\text{TAMLEC}}$ &  0.841  & 0.803  & 0.797  & 0.810  & 0.810  & 0.824\\
		&  TAMLEC &  \textbf{0.866}  & \textbf{0.835}  & \textbf{0.837}  & \textbf{0.843}  & \textbf{0.850}  & \textbf{0.873}\\
		\midrule
		\parbox[t]{2mm}{\multirow{8}{*}{\rotatebox[origin=c]{90}{ EURLex}}} &  MATCH &  0.852  & 0.886  & 0.847  & 0.891  & 0.860  & 0.851\\
		&  XML-CNN &  0.869  & 0.917  & 0.892  & 0.923  & 0.907  & 0.907\\
		&  Att.XML &  0.870  & 0.893  & 0.861  & 0.897  & 0.871  & 0.860\\
		&  HECTOR &  0.914  & 0.921  & 0.896  & 0.926  & 0.908  & 0.877\\
		&  MAML-T &  0.644  & 0.755  & 0.767  & 0.768  & 0.780  & 0.766\\
		&  $\sqrt{\text{TAMLEC}}$ &  0.910  & 0.906  & 0.863  & 0.911  & 0.876  & 0.825\\
		&  TAMLEC &  \textbf{0.945}  & \textbf{0.954}  & \textbf{0.925}  & \textbf{0.958}  & \textbf{0.936}  & \textbf{0.911}\\
		\bottomrule
	\end{tabular}

	\caption{Performance comparison of \algoName and other competing methods on the XMLCo experiment on the MAG-CS, PubMed and EURLex datasets. The best values for each combination of metric and dataset are written in bold.}
	\label{tab:exp_ranking_metrics}
\end{table}

\noindent\textbf{Evaluation and Metrics.}
Throughout our experiments, we evaluate the different methods using their ranked label prediction $\mathcal{R}$, which list the labels by decreasing order of predicted probability.
 We compare the different ranked label prediction using two metrics: Precision at $k$ ($P@k$) and  Normalized Discounted Cumulative Gain at $k$ ($NDCG@k$). 
Let $y_n = 1$ if $\mathcal{R}(n)$ is correct and $0$ otherwise.
In other words, $y_n$ is a boolean variable that indicates whether the n-th element of $\mathcal{R}$ belongs to the document.
With these notations, $P@k$ is defined as: 
\begin{equation*}
	P@k = \frac{1}{k} \sum_{n=1}^{k} y_n,
\end{equation*}
i.e. the average number of correctly predicted labels among the first $k$ elements of $\mathcal{R}$. 
$NDCG@k$ provides a smoother measurement of the quality of the ranking $\mathcal{R}$, by assigning lower weights to failed predictions in the tail of the ranking. Formally, 
\begin{equation*}
	NDCG@k = \frac{\sum_{n=1}^{k} \frac{y_n}{\log(n + 1)}}{\sum_{n=1}^{\min(k, k_y)} \frac{1}{\log(n + 1)}},
\end{equation*}
where  $k_y$ is the number of labels of the documents.
%
%
All metrics (@k) are reported as average on the test set across all the documents that are equipped with at least $k$ labels.

\begin{table*}[!h]
	\vspace{-0.1in}
	
\begin{tabular}{l c| ccc |ccc | ccc | ccc}
	\toprule
	\multirow{2}{*}{} & \multirow{2}{*}{\textbf{Algorithm}} & \multicolumn{3}{c}{\textbf{Global Precision}} & \multicolumn{3}{c}{\textbf{Global NDCG}} & \multicolumn{3}{c}{\textbf{NT Precision}} & \multicolumn{3}{c}{\textbf{NT NDCG}}\\
	\cline{3-14}
	& & @1 & @2 & @3 & @2 & @3 & @4 & @1 & @2 & @3 & @2 & @3 & @4\\
	\midrule
	\parbox[t]{2mm}{\multirow{8}{*}{\rotatebox[origin=c]{90}{ MAG-CS}}} &  MATCH &  0.606 & 0.566 & 0.550 & 0.584 & 0.583 & 0.587 & 0.830 & 0.718 & 0.720 & 0.746 & 0.762 & 0.813\\
	&  XML-CNN &  0.335 & 0.331 & 0.343 & 0.348 & 0.375 & 0.409 & 0.768 & 0.662 & 0.712 & 0.704 & 0.772 & 0.798\\
	&  AttentionXML &  0.662 & 0.645 & 0.646 & 0.662 & 0.680 & 0.700 & 0.711 & 0.614 & \textbf{0.742} & 0.652 & \textbf{0.805} & 0.730\\
	&  HECTOR &  0.035 & 0.022 & 0.016 & 0.025 & 0.020 & 0.017 & \textbf{0.973} & \textbf{0.852} & 0.742 & \textbf{0.881} & 0.796 & \textbf{0.889}\\
	&  MAML-T &  0.320 & 0.317 & 0.287 & 0.355 & 0.353 & 0.369 & 0.715 & 0.667 & 0.636 & 0.695 & 0.738 & 0.637\\
	&  $\sqrt{\text{TAMLEC}}$ &  0.830 & 0.758 & 0.732 & 0.782 & 0.768 & 0.763 & 0.935 & 0.608 & 0.705 & 0.677 & 0.746 & 0.851\\
	&  TAMLEC &  \textbf{0.868} & \textbf{0.799} & \textbf{0.773} & \textbf{0.825} & \textbf{0.812} & \textbf{0.813} & 0.959 & 0.638 & 0.712 & 0.703 & 0.750 & 0.822\\
	\midrule
	\parbox[t]{2mm}{\multirow{8}{*}{\rotatebox[origin=c]{90}{ PUBMed}}} &  MATCH &  0.813 & 0.801 & 0.774 & 0.802 & 0.784 & 0.783 & 0.970 & 0.830 & 0.766 & 0.859 & 0.806 & 0.763\\
	&  XML-CNN &  0.809 & 0.807 & 0.790 & 0.808 & 0.799 & 0.811 & \textbf{1.000} & 0.808 & 0.733 & 0.851 & 0.790 & 0.708\\
	&  AttentionXML &  0.829 & 0.824 & 0.819 & 0.826 & 0.825 & 0.836 & 0.985 & 0.862 & 0.829 & 0.891 & 0.864 & 0.833\\
	&  HECTOR &  0.022 & 0.022 & 0.023 & 0.022 & 0.023 & 0.023 & 1.000 & \textbf{0.937} & \textbf{0.904} & \textbf{0.951} & \textbf{0.926} & \textbf{0.915}\\
	&  MAML-T &  0.546 & 0.517 & 0.504 & 0.526 & 0.531 & 0.607 & 1.000 & 0.701 & 0.602 & 0.769 & 0.685 & 0.607\\
	&  $\sqrt{\text{TAMLEC}}$ &  0.827 & 0.787 & 0.780 & 0.795 & 0.794 & 0.808 & 0.970 & 0.897 & 0.803 & 0.911 & 0.839 & 0.857\\
	&  TAMLEC &  \textbf{0.864} & \textbf{0.832} & \textbf{0.833} & \textbf{0.840} & \textbf{0.847} & \textbf{0.866} & 1.000 & 0.903 & 0.876 & 0.925 & 0.901 & 0.907\\
	\midrule
	\parbox[t]{2mm}{\multirow{8}{*}{\rotatebox[origin=c]{90}{ EURLex}}} &  MATCH &  0.512 & 0.501 & 0.484 & 0.504 & 0.494 & 0.482 & 0.928 & 0.894 & 0.885 & 0.900 & 0.899 & 0.803\\
	&  XML-CNN &  0.335 & 0.356 & 0.367 & 0.352 & 0.364 & 0.372 & 0.946 & 0.905 & 0.926 & 0.915 & 0.947 & \textbf{0.936}\\
	&  AttentionXML &  0.519 & 0.478 & 0.446 & 0.488 & 0.467 & 0.451 & 0.968 & 0.928 & 0.920 & 0.938 & 0.942 & 0.821\\
	&  HECTOR &  0.037 & 0.029 & 0.033 & 0.031 & 0.034 & 0.038 & \textbf{0.968} & 0.957 & 0.951 & 0.960 & 0.964 & 0.849\\
	&  MAML-T &  0.654 & 0.742 & 0.744 & 0.758 & 0.769 & 0.716 & 0.914 & 0.876 & 0.858 & 0.885 & 0.900 & 0.857\\
	&  $\sqrt{\text{TAMLEC}}$ &  0.913 & 0.913 & 0.877 & 0.918 & 0.888 & 0.836 & 0.954 & 0.941 & 0.897 & 0.943 & 0.913 & 0.764\\
	&  TAMLEC &  \textbf{0.944} & \textbf{0.955} & \textbf{0.914} & \textbf{0.959} & \textbf{0.928} & \textbf{0.895} & 0.968 & \textbf{0.966} & \textbf{0.963} & \textbf{0.967} & \textbf{0.974} & 0.869\\
	\bottomrule
\end{tabular}

\caption{ Performance comparison on the Few-Shot experiment on the MAG-CS, PubMed and EURLex datasets. NT (resp. Global) indicates a metric computed on the new task (resp. all the tasks, including the new task). The best values for each combination of metric and dataset are written in bold.}
\label{tab:exp_fewshots_metrics}
\end{table*}


\subsection{Label Completion} \label{subsec:xp_XMLCO}

\noindent \textbf{Experimental design.} For the XMLCo task, each document is associated to an incomplete set of labels that must be enriched. 
In our experiments, we simulate an XMLCo problem by removing all the labels from each document except the ones associated with the most general concepts in the taxonomy.
The motivation behind this choice is twofold: first, the most general labels are the easiest to obtain -- for instance, the venue at which a scientific paper is published often suffices to deduce its field of research. 
Moreover, they are the one most commonly present in a dataset \cite{DBLP:journals/cbm/RomeroNFRV23,ostapuk2024follow}.
Second, due to our hierarchical assumption, more general labels can easily be deduced from more specific ones.

During the task of label completion, \algoName is able to map each document to its related TAT(s) accurately using the general concepts contained partial set of label.
To use the different baselines for XMLCo, we modify them as follows: 
%
we run a normal inference step and then skip model predictions of labels 
that already belongs to the document. 
%
%
Regarding Hector \cite{ostapuk2024follow}, it is important to note that the taxonomies considered in this experiment cannot be fully encoded by a tree -- a requirement for Hector predictions.
Thus, we modify the taxonomies provided to hector by removing the minimum amount of relations possible to reduce each weak-semilattice to a tree, in line with the experiments of its paper.

\smallskip

\noindent \textbf{Results.} Table \ref{tab:exp_ranking_metrics} reports the results of the XMLCo experiments.
First, \algoName performs significantly better than the other XML methods (MATCH, XML-CNN, AttentionXML and HECTOR) across almost all metrics and datasets.
The only exception is Precision@3, where AttentionXML achieves the best performance on both the MAG-CS dataset, but only by a very narrow margin ($0.002$).
The advantage of \algoName is particularly visible on EURLex, where it outperforms the other baselines by a wide margin. 
This can be explained by the fact that EURLex TATs decomposition contains more than 100 TATs, significantly more than the other datasets, maximizing the benefits of \algoName TATs' approach.
%
Second, MAML-T achieves poor performance on the different XMLCo metrics.
These results can be explained by the fact that XML is a notoriously difficult problem, for which MAML (the base meta-algorithm) was not designed. 
%
%
Finally, it is important to note \algoName performs better than \algoAblation for both Precision and NDCG.
This difference highlights the fact that all the components of \algoName are necessary for its high performance.


\subsection{Few Shots XML}\label{subsec:xp_FewShots}

\noindent \textbf{Experimental design.} In this experiment, we aim at evaluating the Few-shot XML potential of \algoName, when compared to all the XMLC baseline  and the  Few-shots algorithm MAML-T 
in particular.
To simulate a few-shot setting, we proceed as follows: during training, a TAT $T_i$ is withheld from the training set, by removing all the labels of $T_i$ as well as all the documents that are solely equipped with labels of $T_i.$ 
After being trained on this large training set, the different algorithms are presented with the new task for fine-tuning: this is performed either by training the models for a few epochs  on the new task (MATCH, XML-CNN, AttentionXML, HECTOR, \algoName), or by using the algorithm specific few-shot training process (MAML-T).
For \algoName, only the parameters associated to the generator specific to this new task are trained, resulting in an extremely fast second training. 
%


\smallskip

\noindent \textbf{Results.} 
Table \ref{tab:exp_fewshots_metrics} reports for every combination of algorithm and dataset, the metrics on both the new task (NT) and the complete dataset (Global).
First, \algoName performance is  much higher than other methods on the Global metrics.
In fact, its performance are barely lower than in the regular experiment (Table \ref{tab:exp_ranking_metrics}), where all the data are present during preliminary training. 
This indicate that \algoName is able to adapt seamlessly to the appearance of a completely new task after the initial training, as most of the adaptation is done using the dedicated generator bloc.
The other XMLC methods exhibit significantly worse global performance, as they try to adapt to the new task. 
The worsening is particularly pronounced for HECTOR, which appear to only predict labels relevant to the new task in our experiments, resulting in the extremely poor global performance.
Conversely, MAML-T global performance only varied by a small amount, which is illustrative of their Few-shots design. 

As the different methods were trained specifically on the new task, most algortihms were able to achieve very strong NT metrics. 
While \algoName does not always achieve the best results, it is important to note that its metrics are always close to the best performing method on the new task. 
Furthermore, and as discussed above, the better performance of other methods such as HECTOR on the new task come at a drastic cost to their global results.
Thus, we argue that \algoName achieves the best trade-off between adapting to the new task and retaining performance on the entire problem.

%
%
%

\section{Conclusion and Future Work}
\label{sec:conclu}

In this paper, we introduced a new algorithm, \algoName, that tackles XMLCo problems by dividing the associated taxonomy into Taxonomy-Aware Tasks (TATs), where each task is adapted to the path structure of the taxonomy.
By leveraging ideas from parallel feature sharing and multi-tasks learning, \algoName uses the TATs decomposition to achieve substantially better performance on label completion and few shots XML problems across multiple datasets.
Future works include the study of new types of TATs decomposition, where some TATs may be more specific (i.e. staring with more specific concepts) than others, as well as developing new Parallel Feature sharing methods to better embed the TATs into the weight sharing.

\flushend
\balance
\bibliographystyle{ACM-Reference-Format}
\bibliography{sample-base}


\end{document}